\def\year{2020}\relax
\documentclass[letterpaper]{article} 
\usepackage{aaai20}  
\usepackage{times}  
\usepackage{helvet} 
\usepackage{courier}  
\usepackage[hyphens]{url}  
\usepackage{graphicx} 
\usepackage{graphicx}  
\usepackage{comment}
\usepackage{amsmath,amssymb} 
\usepackage{color}
\usepackage{multirow}
\usepackage{booktabs}
\usepackage{url}            

\usepackage[switch]{lineno}  %

\title{Improving Adversarial Robustness via Probabilistically Compact Loss with Logit Constraints}

\author{Xin Li\thanks{These authors contributed equally.},Xiangrui Li\footnotemark[1], Deng Pan\footnotemark[1], Dongxiao Zhu\thanks{Corresponding author.}\\
Department of Computer Science, Wayne State University, Detroit, MI 48202 \\
\{xiangruili, xinlee, pan.deng, dzhu\}@wayne.edu}

\begin{document}

\maketitle

\begin{abstract}
Convolutional neural networks (CNNs) have achieved state-of-the-art performance on various tasks in computer vision. However, recent studies demonstrate that these models are vulnerable to carefully crafted adversarial samples and suffer from a significant performance drop when predicting them. Many methods have been proposed to improve adversarial robustness (e.g., adversarial training and new loss functions to learn adversarially robust feature representations). Here we offer a unique insight into the predictive behavior of CNNs that they tend to misclassify adversarial samples into the most probable false classes. This inspires us to propose a new Probabilistically Compact (PC) loss with logit constraints which can be used as a drop-in replacement for cross-entropy (CE) loss to improve CNN's adversarial robustness. Specifically, PC loss enlarges the probability gaps between true class and false classes meanwhile the logit constraints prevent the gaps from being melted by a small perturbation. We extensively compare our method with the state-of-the-art using large scale datasets under both white-box and black-box attacks to demonstrate its effectiveness. The source codes are available from the following url: \url{https://github.com/xinli0928/PC-LC}.

\end{abstract}

\section{Introduction}

Convolutional neural networks (CNNs) have achieved significant progress for various challenging tasks in computer vision, including image classification \cite{DBLP:conf/aaai/LiLPZ20}, semantic segmentation \cite{he2017mask}, and image generation \cite{goodfellow2014generative}. Despite their success, CNNs are highly vulnerable to adversarial samples \cite{szegedy2013intriguing}. With imperceptibly small perturbation added to a clean image, adversarial samples can drastically change models' prediction, resulting in a significant drop in CNN's predictive performance. This phenomenon poses a serious threat to security-critical applications of deep learning, such as autonomous driving \cite{al2017deep}, surveillance system \cite{sreenu2019intelligent}, and medical imaging system \cite{DBLP:conf/isbi/LiZ20}. Furthermore, studies have shown that adversarial robustness is also a key property to obtain human interpretation in computer vision and other application fields \cite{DBLP:journals/corr/abs-1912-03430,DBLP:conf/ijcai/PanLLZ20}. Therefore, improving models' adversarial robustness is critical to build trustworthy Artificial Intelligence systems to prevent unforeseen hazardous situations.

To improve CNN's adversarial robustness, many methods have been proposed. One strategy is to modify the inputs during inference time via noise removal \cite{hendrycks2016early,meng2017magnet}, super-resolution \cite{mustafa2019image} and JPEG compression \cite{dziugaite2016study,das2017keeping} to diminish the impact of perturbation, but can be easily evaded by strong attacks \cite{athalye2018obfuscated}. Another type of strategy \cite{tramer2017ensemble,kurakin2016adversarial,sinha2017certifiable,zhang2019you,shafahi2019adversarial} is based on adversarial training \cite{goodfellow2014explaining} that can effectively increase the model's robustness by utilizing crafted adversarial examples as data augmentation. However, it is computationally expensive and compromise model classification performance on clean images \cite{tsipras2018robustness}. Other than modifying data, some techniques directly enhance model robustness by altering network architectures \cite{taghanaki2019kernelized,mustafa2019adversarial}, or constructing ensembles of networks \cite{tramer2017ensemble,pang2019improving}. However, they require additional processes and are not flexible to be adopted to other models. 

While previous works have successfully improved CNN robustness against adversarial attacks, the connection of CNN predictions between adversarial and clean samples is not known. In this paper, we investigate this connection in the typical setting when CNNs are trained with the cross-entropy (CE) loss. When an adversarial sample successfully fools the trained CNN with small perturbations, it tends be to misclassified into the first several most probable false classes when predicting the original clean sample, i.e., the classes with larger predicted probabilities. This consistent pattern of CNN's predictive behaviors is intuitive and potentially implies a deeper connection between the CNN feature learning and its adversarial robustness. 

\begin{figure}[t]
	\centering
	\includegraphics[width=\columnwidth]{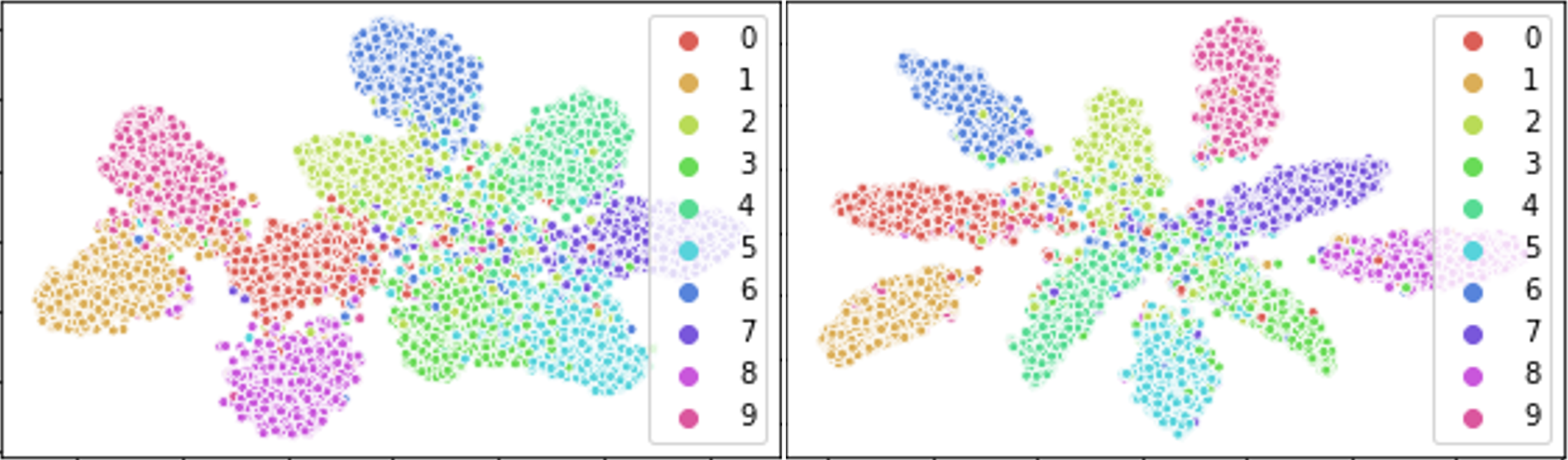}
	\caption{T-SNE visualization of the penultimate layer of ResNet-56 trained with CE loss (left) and PC loss (right) on CIFAR-10.}
	\label{fig:vis}
\end{figure}

The tendency of misclassifying adversarial samples enlightens us that the adversarial robustness of CNN can be benefited from the training that focuses on the differentiation between sample's true class and its first several most probable false classes. Hence, in this paper, we propose a novel training objective, termed as Probabilistically Compact (PC) loss with logit constraints, which can improve adversarial robustness and achieve comparable classification accuracy without extra training procedure and computational burden. 

Unlike CE loss which focuses only on maximizing the output probability of the true class, PC loss aims at maximizing probability gaps between the true class and the most probable false classes. Meanwhile, the logit constraints suppress logit which ensures that the gaps is not only large, but also difficult to be crossed. Consequently, this formulation helps to widen the gaps between different classes in feature space, and ensures that it is difficult for an adversary to fool the trained model with small perturbations. We demonstrate the synergistic effect of the gaps at both probability and feature levels using benchmark datasets. For example, the average probability gaps between the true class and the most probable false class of ResNet-56 on CIFAR-100 test data are 0.527 (CE loss) vs. 0.558 (PC loss), respectively. And for Tiny ImageNet test data the gaps become 0.131 (CE loss) vs. 0.231 (PC loss). These results demonstrate that our PC loss can directly enlarge the probability gap of prediction and the effect is more pronounced for more challenging dataset (Tiny ImageNet). As shown in Figure \ref{fig:vis}, ResNet-56 trained with our PC loss has clear margin boundaries and samples of each classes are evenly distributed around the center with a minimal overlap on CIFAR-10 test data. 

Our main contributions are summarized as follows: (1) We offer an unique insight into the predictive behavior of CNN on adversarial samples that the former tends to misclassify the latter into the first several most probable classes. (2) We formulate the problem by proposing a new loss function, i.e., PC loss with logit constraints to improve CNN's adversarial robustness, where these two components are systematically integrated and simultaneously optimized during training process. (3) Our PC loss can be used as a drop-in replacement of the CE loss to supervise CNN training without extra procedure nor additional computational burden for improving adversarial robustness. Experimental results show that when trained with our method, CNNs can achieve significantly improved robustness against adversarial samples without compromising performance on predicting clean samples.

\section{Related Work}
For generating adversarial samples, various computational methods have been developed. Fast Gradient Sign Method (FGSM) \cite{goodfellow2014explaining} and its variant Basic Iterative Method (BIM) efficiently generates adversarial samples by perturbing the pixels according to the gradient of the loss function \cite{kurakin2016adversarial}. Projected Gradient Descent (PGD) \cite{madry2017towards} introduces a random starting point at each iteration in FGSM within a specified $l_\infty$ norm-ball to enhance the attack effect. Momentum iterative method (MIM) \cite{dong2018boosting} uses momentum to help iterative gradient-based methods to avoid sticking into local maximum thus further boosts their attacking performance. As an optimization-based attack, the Carlini and Wagner (C\&W) \cite{carlini2017towards} approach uses binary search mechanism to find the minimal perturbation for a successful attack. SPSA \cite{uesato2018adversarial} is a gradient-free method which approximates gradient to generate attacks and defeats many defenses. It outperforms gradient-based attacks when the loss surface is hard to optimize.

To counter adversarial attack and enhance model robustness, various defensive techniques have been proposed. Among these proposed defenses, one line of those approaches \cite{kurakin2016adversarial,sinha2017certifiable,zhang2019you,shafahi2019adversarial} are based on adversarial training \cite{goodfellow2014explaining} and can achieve effective robustness against different adversarial attacks, where the training dataset is augmented with adversarial samples. However, these methods have trade-off between accuracy on clean images and adversarial robustness \cite{tsipras2018robustness} and are computationally expensive in adversarial samples generation \cite{zhang2019you}. To reduce the computational burden, Shafahi et al. \cite{shafahi2019adversarial} propose a training algorithm, which improves the efficiency of adversarial training by updating both model parameters and image perturbation in one backward pass.

Another line of defending strategy against adversaries, other than augmenting the training dataset, is to learn feature representations with adversarial robustness by using model ensembles or altering network architectures \cite{taghanaki2019kernelized,mustafa2019adversarial,tramer2017ensemble,pang2019improving}. For example, \cite{taghanaki2019kernelized} augment CNNs with the radial basis function kernel to further transform features via kernel trick to improve the class separability in feature space and reduce the effect of perturbation. \cite{mustafa2019adversarial} propose a prototype objective function, together with multi-level deep supervision. Their method ensures the feature space separation between classes and shows significant improvement of robustness. \cite{pang2019improving} obtain a strong ensemble defense by introducing a new regularizer to encourage diversity among models within the ensemble system, which encourages the feature representation from the same class to be close. Although these approaches avoid high computationally cost of adversarial training, they have to modify the network or require extra training process, which limits their flexibility to be adapted to different tasks.

\begin{figure*}[t]
\centering
\includegraphics[width=\linewidth]{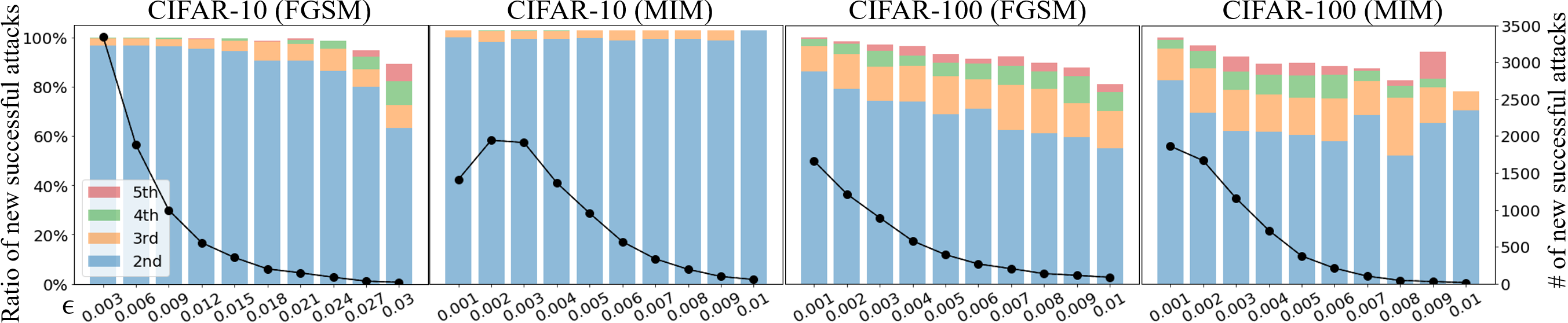}
\caption{Empirical investigation on the predictive behavior of CNN on adversarial samples from CIFAR-10 and CIFAR-100. The line (black, right y-axis) represents the number of increased successful attacks when $\boldsymbol{\epsilon}$ is increased from its previous grid value. Each bar (left y-axis) represents the percentage of misclassification for the increased successful attacks, measuring number of adversarial samples are misclassified into the 2nd, 3rd, 4th and 5th most probable classes. FGSM and MIM are attack methods.}
\label{fig:motivation}
\end{figure*}

More efficient approaches are designing new loss functions to improve model adversarial robustness. By explicitly imposing regularization on latent features, CNNs are encouraged to learn feature representations with more inter-class separability and intra-class compactness \cite{pang2019rethinking,elsayed2018large,mustafa2019adversarial}. For example, \cite{pang2019rethinking} propose a Max-Mahalanobis center (MMC) loss to learn discriminative features. They first calculate Max-Mahalanobis \cite{pang2018max} centers for each class and then encourage the features to gather around the centers using Center Loss \cite{wen2016discriminative}. However, the assumption of geometrical compactness for latent features (in terms of Euclidean distance or $L_2$-norm) may not hold due to inherent intra-class variations in the data and usually requires suitable assumptions on distribution of the latent features. Differently, our PC loss with logit constraints is motivated by the predictive behavior of CNN on adversarial samples from the probability perspective, which avoids this issue by learning probabilistically compact features without geometric assumptions. The work that is closest to ours is \cite{chen2019complement} that encourages the predicted probabilities of false classes to be equally distributed, whereas our PC loss directly enlarges the gap of probabilities between true class and the first several most probable false classes.

\section{Proposed Method} \label{sec:PCLoss}


\noindent\textbf{Notation} Let $D = {(\boldsymbol{x}_i, y_i)}_{i=1}^{N}$ be the set of training samples of size $N$, where $\boldsymbol{x}_i \in \textbf{R}^p$ is the $p$-dimensional feature vector and $y_i = k (k=1,\cdots, K)$ is the true class label, and $S_k= \{(\boldsymbol{x}_i, y_i): y_i=k\}$ the subset of $D$ for the $k$-th class. The bold $\boldsymbol{y}_i=(y_i^1,\cdots,y_i^K)$ is used to represent the one-hot encoding for $y_i$: $y_i^k=1$ if $y_i = k$, $0$ otherwise.

\noindent\textbf{Cross-entropy (CE) loss} Assume that CNN's output layer, after convolutional layers, is a fully connected layer of $K$ neurons with bias terms, then the predicted probability for sample $\boldsymbol{x}$ being classified into $k$-th class is calculated using the softmax activation (the $k$-th logit $a_k= \boldsymbol{W}_k\boldsymbol{h}_{\boldsymbol{x}}+b_k$):
\begin{equation} \label{eq:prob}
f_k(\boldsymbol{x}) = p(y =k|\boldsymbol{x})  = \frac{\exp(a_k)}{\sum_{j=1}^{K}\exp(a_j)} \hspace{3mm}(k=1,\cdots, K),
\end{equation}
where $\boldsymbol{h}_{\boldsymbol{x}}$ is the feature representation of $\boldsymbol{x}$, $\boldsymbol{W}_k$ and $b_k$ are parameters of the $k$-th neuron in the output layer. Then CE loss, which is equivalent to the maximum likelihood approach, is given as follows:
\begin{equation} \label{eq:RSML}
L(\boldsymbol{\theta}) = -\sum_{k=1}^{K} \sum_{i_k\in S_k}  \log f_k(\boldsymbol{\theta};\boldsymbol{x}_{i_k}) ,
\end{equation}
where $\boldsymbol{\theta}$ is the vector of trainable model parameters.

\subsection{Motivation: Predictive Behavior of CNN on Adversarial Samples}\label{subsec:motivation}
Previous studies have shown that when trained to optimum, \textit{i.e.} $\boldsymbol{\theta}^* = \arg\min_{\boldsymbol{\theta}}L(\boldsymbol{\theta})$, CNNs can misclassify adversarial samples that are only slightly different from the original clean samples. This vulnerability has recently inspired many methods for generating adversarial samples (attack), defending adversarial attacks and detecting adversarial samples. Here, we take a different perspective on the attacks and empirically investigate if there is a systematic tendency on how CNNs misclassify adversarial samples.

Specifically, for a testing (clean) sample $(\boldsymbol{x},y)$, the (untargeted) attack seeks a small perturbation $\boldsymbol{\epsilon}$ that leads to the misclassification of $\boldsymbol{x}$ when the perturbation is added to $\boldsymbol{x}$:
\begin{equation} \label{prob:minperturb}
   \min_{\boldsymbol{\epsilon}} ||\boldsymbol{\epsilon}||_p, \text{s.t. } y'= \arg\max_k f_k(\boldsymbol{x}+\boldsymbol{\epsilon}) \text{ and } y\neq y',
\end{equation}
where $||\cdot||_p$ is the norm such as $L_1$, $L_2$ and $L_\infty$. When Eq. (\ref{prob:minperturb}) is optimized and the attack succeeds, a natural question to ask is \textit{``are there any connections between $y'$ and $y$ for the trained CNN misclassifying $\boldsymbol{x+\epsilon}$ into class $y'$?"} Note that $y=\arg\max_k f_k(\boldsymbol{x})$.

Intuitively, we could expect that $y'$ is likely to be the most probable class except the true class label $y$, \textit{i.e}, the class corresponding to the 2nd largest value of CNN predicted probabilities. This conjectures that solving Eq. (\ref{prob:minperturb}) is equivalent to solve
\begin{equation} \label{prob:minperturb2nd}
  \min_{\boldsymbol{\epsilon}} ||\boldsymbol{\epsilon}||_p, \text{s.t.}\arg\max_k f_k(\boldsymbol{x}+\boldsymbol{\epsilon})
  =\arg_{\#2}\max_k f_k(\boldsymbol{x}),
\end{equation}
where $\arg_{\#2}\max$ represents the operation of taking the 2nd largest value\footnote{We may relax it to the first several most probable classes such as 3rd and 4th.}.

Here we provide an analysis as our motivation behind this conjecture. Assuming that the CNN is Lipschitz continuous \cite{fazlyab2019efficient}, then we have the inequality: 
\begin{equation}\label{eq:lipschitz}
   ||\boldsymbol{f}(\boldsymbol{x}+\boldsymbol{\epsilon}) - \boldsymbol{f}(\boldsymbol{x})||_p \leq l||(\boldsymbol{x+\epsilon}) - \boldsymbol{x}||_p = l||\boldsymbol{\epsilon}||_p, 
\end{equation}
where $l$ is the Lipschitz constant and $\boldsymbol{f(\cdot)} = (f_1(\cdot),\cdots,f_K(\cdot))$. The Lipschitz continuity implies that the change of the ouput is bounded by the change of the input, which is the small perturbation in adversarial attacks. To misclassify $\boldsymbol{x+\epsilon}$, the possible minimal value of the LHS in Eq. (\ref{eq:lipschitz}) is to seek an $\boldsymbol{\epsilon}$ such that $f_{j}(\boldsymbol{x+\epsilon}) \geq f_y(\boldsymbol{x+\epsilon})$, where $y$ is true class and $j$ is 2nd most probable class. To see this, consider the following two cases:
\begin{itemize}
    \item Case 1. $f_y(\boldsymbol{x+\epsilon})\geq f_y(\boldsymbol{x})$. To misclassify $\boldsymbol{x+\epsilon}$, the possible minimal value $||\boldsymbol{f}(\boldsymbol{x}+\boldsymbol{\epsilon}) - \boldsymbol{f}(\boldsymbol{x})||_p$ is to reduce $f_{k'}(\boldsymbol{x})$ $(k'\neq y, j)$ to compensate $f_{j}(\boldsymbol{x+\epsilon})$ so that $f_{j}(\boldsymbol{x+\epsilon}) \geq f_y(\boldsymbol{x+\epsilon})$.
    \item Case 2. $f_y(\boldsymbol{x+\epsilon})< f_y(\boldsymbol{x})$. The possible minimal value is that $f_y(\boldsymbol{x+\epsilon}) = f_y(\boldsymbol{x})-(\frac{f_y(\boldsymbol{x}) - f_{j}(\boldsymbol{x})}{2})$, $f_{j}(\boldsymbol{x+\epsilon}) = f_{j}(\boldsymbol{x}) + (\frac{f_y(\boldsymbol{x}) - f_{j}(\boldsymbol{x})}{2})$ and all other $f_{k'}(\boldsymbol{x})$ $(k'\neq y, j)$ remain unchanged.
\end{itemize}
Those two cases may not be achievable in practice, but provide a lower bound on $||\boldsymbol{f}(\boldsymbol{x}+\boldsymbol{\epsilon}) - \boldsymbol{f}(\boldsymbol{x})||_p$. The same analysis can be further relaxed to the 3rd and 4th most probable classes. Observing Eq. (\ref{eq:lipschitz}), solving Eq. (\ref{prob:minperturb}) provides an upper-bound for the LHS of Eq. (\ref{eq:lipschitz}). With Lipschitz continuity, we hence conjecture that CNN tends to misclassify adversarial samples into classes that have large predicted probabilities when predicting the original clean samples.


To verify our conjecture, we perform an empirical study on CIFAR-10 and CIFAR-100 datasets. FGSM and MIM are used as the adversarial attack algorithms and generate adversarial samples for the standard testing data of CIFAR-10 and CIFAR-100. We do not solve Eq. (\ref{prob:minperturb}) for each testing samples as it is computationally expensive for the test data of size 10,000. Instead, we take a fine grid of perturbation values and summarize the misclassification of newly successful attacks when the perturbation $\boldsymbol{\epsilon}$ is increased from $\boldsymbol{\epsilon}_m$ to $\boldsymbol{\epsilon}_{m+1} = \boldsymbol{\epsilon}_m +\Delta$ ($\Delta$ is the value of increment). Figure~\ref{fig:motivation} displays the summary of the misclassification results. From the figure, we can see that for CIFAR-10, every time the perturbation is increased, the newly successful attacks are mostly misclassified into the 2nd most probable class for the clean samples. For CIFAR-100, the misclassification follows a similar trend considering it has 100 classes. We also notice that as the perturbation gets larger in FGSM, more newly successful attacks are classified into the 3rd, 4th and 5th most probable classes of predicting clean samples. A possible reason is that the large perturbation results in overshoot in the misclassification as the difference between 2nd most probable class and 3rd most probable class is small when a clean sample needs large perturbation to be adversarial. Different from FGSM, as perturbation increases, MIM always maintains a high percentage of classifying newly successful adversarial samples into the 2nd most probable class of predicting clean samples for CIFAR-10, due to its iterative procedure in generating adversarial attacks. Overall, Figure~\ref{fig:motivation} empirically agrees with our analysis that motivates our proposed PC loss.

\subsection{Probabilistically Compact Loss}\label{subsec:pcl}
The predictive behavior of CNN on adversarial samples in the last section inspires us that to improve model robustness to adversaries, CNN needs to focus on the differentiation between the true class and the first several most probable classes. In terms of predicted probability, CNN robustness is benefited from the large gap between true class $f_y(\boldsymbol{x})$ and false class $f_{y'}(\boldsymbol{x})$ $({y'}\neq y)$. Indeed, \cite{neyshabur2017exploring} shows that the gap $f_y(\boldsymbol{x})-\max_{y'} f_{y'}(\boldsymbol{x})$ can be used to measure the generalizability of deep neural networks.

With the aforementioned motivation, we propose the PC loss  to improve CNN's adversarial robustness as follows:
\begin{equation}\label{eq:pcl}
L_{pc}(\boldsymbol{\boldsymbol{\theta}}) =\frac{1}{N}\sum_{{y'}\neq y_i, i\in D}\max(0, f_{y'}(\boldsymbol{x}_{i}) +\xi -f_{y_i}(\boldsymbol{x}_{i})),
\end{equation}
where $N$ is the number of training samples, $\xi >0$ is the probability margin treated as hyperparameter. Here, we include all non-target classes in the formulation and penalize any classes for each training sample that violate the margin requirement for two considerations: (1) if one of the most probable classes satisfies the margin requirement, all less probable classes will automatically satisfy this requirement and hence have no effect in PC loss; (2) since the first several most probable classes are unknown and can change during the training process, it is necessary to maintain the margin requirement for all classes.

Compared with previous works \cite{mustafa2019adversarial,pang2019rethinking,taghanaki2019kernelized} that improve adversarial robustness via explicitly learning features with large intra-class compactness, PC loss avoids assumptions on the feature space. Instead, PC loss only encourages the feature learning that leads to probabilistic intra-class compactness by imposing a probability margin $\xi$. 

In training CNN with PC loss, the latter is differentiable and hence can be optimized with stochastic gradient descent. The gradient of PC loss can be calculated as (w.r.t. logit $a_y$)
\begin{equation} \label{eq:mpcDir}
\frac{\partial (f_{y'} + \xi -f_y)}{\partial a_y} = -f_y (1-f_y+f_{y'}),
\end{equation}
where the gradient is computed for the softmax function. For other logits ($a_{y'}$), the gradients can be similarly computed.

\subsection{The Logit Constraints}\label{subsec:logit}

In last section we introduce PC loss to enhance adversarial robustness by enlarging the probability gaps. In this section, we further propose logit constraints as a complement of PC loss, which suppress logit to ensure that the gaps are not only large, but also difficult to be crossed. Here we explain the necessity of using both parts together. We use $|\cdot|$ to denote $||\cdot||_2$ for simplicity. The conclusion can be extended to other $L_p$ norms.  Assume there is a clean image $\boldsymbol{x}$, and a corrupted image $\boldsymbol{x}+\boldsymbol{\epsilon}$ attacked by some adversarial algorithm, where $|\boldsymbol{\epsilon}|<\tau$, with $\tau > 0$ be a small constant. 

We take the log probability for simplicity, i.e., PC loss is equivalently enlarging $\log f_y(\boldsymbol{x})-\log f_j(\boldsymbol{x})$ between true class $y$ and most probable false class $j$. Given a perturbation $\boldsymbol{\epsilon}$, the corresponding log probabilities can be estimated via first order approximation 
\begin{equation}
    \log f_y(\boldsymbol{x} + \boldsymbol{\epsilon}) = \log f_y(\boldsymbol{x}) + \boldsymbol{\epsilon}\cdot \nabla_{\boldsymbol{x}} \log f_y(\boldsymbol{x}),
\end{equation}
$\log f_j(\boldsymbol{x}+\boldsymbol{\epsilon})$ can also be approximated in the same manner. To prevent $\log f_y(\boldsymbol{x} + \boldsymbol{\epsilon}) -  \log f_j(\boldsymbol{x} + \boldsymbol{\epsilon}) < 0$ (i.e., false prediction with perturbation $\boldsymbol{\epsilon}$), we should solve
$   \min_{\boldsymbol{\theta}} \boldsymbol{\epsilon}\cdot (\nabla_{\boldsymbol{x}} \log f_j(\boldsymbol{x}) - \nabla_{\boldsymbol{x}}\log f_y(\boldsymbol{x})).$

Lets denote vector $\boldsymbol{b} = \nabla_{\boldsymbol{x}} \log f_j(\boldsymbol{x}) - \nabla_{\boldsymbol{x}}\log f_y(\boldsymbol{x})$. As the attackers can always choose the worst $\hat{\boldsymbol{\epsilon}}$ that maximizes $\boldsymbol{\epsilon}\cdot \boldsymbol{b}$ by letting $\hat{\boldsymbol{\epsilon}}$ in the same direction as $\boldsymbol{b}$, i.e. $\boldsymbol{\epsilon}\cdot \boldsymbol{b} \leq |\hat{\boldsymbol{\epsilon}}|\cdot |\boldsymbol{b}|$. Our goal becomes to minimize the upper bound $ |\hat{\boldsymbol{\epsilon}}|\cdot |\boldsymbol{b}|$
\begin{equation}
\begin{split}
    \min_{\boldsymbol{\theta}}  |\hat{\boldsymbol{\epsilon}}||\boldsymbol{b}| &\Longrightarrow \min_{\boldsymbol{\theta}}|\boldsymbol{b}| \\&\Longrightarrow \min_{\boldsymbol{\theta}}  |\nabla_{\boldsymbol{x}} \log f_j(\boldsymbol{x}) - \nabla_{\boldsymbol{x}}\log f_y(\boldsymbol{x})|.
\end{split}
\end{equation}

\begin{table*}[t]
	\begin{center}
		  \resizebox{0.75\textwidth}{!}{
		\begin{tabular}{c|c|cc|cc|cc|c|cc|cc}
			\hline \hline
			\multirow{2}{*}{Attacks} & \multicolumn{1}{c|}{\multirow{2}{*}{Param.}} & \multicolumn{2}{c|}{MNIST}                          & \multicolumn{2}{c|}{KMNIST}                         & \multicolumn{2}{c|}{FMNIST}                         & \multicolumn{1}{c|}{\multirow{2}{*}{Param.}} & \multicolumn{2}{c|}{CIFAR-10}                         & \multicolumn{2}{c}{SVHN}               \\
			& \multicolumn{1}{c|}{}                        & \multicolumn{1}{c}{CE} & \multicolumn{1}{c|}{Ours} & \multicolumn{1}{c}{CE} & \multicolumn{1}{c|}{Ours} & \multicolumn{1}{c}{CE} & \multicolumn{1}{c|}{Ours} & \multicolumn{1}{c|}{}                        & \multicolumn{1}{c}{CE} & \multicolumn{1}{c|}{Ours} & \multicolumn{1}{c}{CE} & Ours          \\ \hline\hline
			Clean                    & -                                          & 99.2                    & 99.2                      & 95.5                    & 95.4                      & 90.1                    & 90.2                      & -                                          & 91.6                    & 91.2                      & 94.9                    & 94.7          \\ \hline
			\multirow{3}{*}{FGSM}    & $ 0.1$                               & 71.5                    & \textbf{80.5}             & 26.2                    & \textbf{62.8}             & 17.5                    & \textbf{58.0}             & $ 0.04$                              & 4.3                    & \textbf{53.1}             & 8.7                    & \textbf{39.5} \\
			& $ 0.2$                               & 51.6                    & \textbf{76.3}             & 1.8                     & \textbf{39.7}             & 9.5                     & \textbf{43.3}             & $ 0.12$                              & 11.7                     & \textbf{30.3}             & 4.5                    & \textbf{24.2} \\
			& $ 0.3$                               & 31.8                    & \textbf{65.0}             & 1.2                     & \textbf{34.1}             & 7.6                     & \textbf{31.8}             & $ 0.2$                               & 11.5                     & \textbf{18.7}              & 2.6                    & \textbf{17.1} \\ \hline
			\multirow{3}{*}{BIM}     & $ 0.1$                               & 52.8                    & \textbf{72.0}             & 55.8                    & \textbf{82.5}             & 0.0                     & \textbf{18.7}             & $ 0.04$                              & 0.0                    & \textbf{29.0}             & 1.3                    & \textbf{26.2} \\
			& $ 0.2$                               & 4.5                     & \textbf{48.6}             & 28.7                    & \textbf{73.3}             & 0.3                     & \textbf{8.4}              & $ 0.12$                              & 0.0                    & \textbf{21.0}             & 0.0                     & \textbf{18.2} \\
			& $ 0.3$                               & 1.5                     & \textbf{39.5}             & 16.8                    & \textbf{60.5}             & 0.0                     & \textbf{6.4}              & $ 0.2$                               & 0.0                     & \textbf{20.0}             & 0.0                     & \textbf{17.6} \\ \hline
			\multirow{3}{*}{PGD}     & $ 0.1$                               & 49.0                    & \textbf{72.3}             & 31.3                    & \textbf{62.4}             & 0.0                     & \textbf{15.7}             & $ 0.04$                              & 0.0                     & \textbf{27.6}             & 0.0                     & \textbf{27.6} \\
			& $ 0.2$                               & 3.3                     & \textbf{50.2}             & 3.9                     & \textbf{39.9}             & 0.0                     & \textbf{7.0}              & $ 0.12$                              & 0.0                     & \textbf{14.6}             & 0.0                     & \textbf{22.0} \\
			& $ 0.3$                               & 0.8                     & \textbf{39.7}             & 2.0                     & \textbf{33.2}             & 0.0                     & \textbf{4.6}              & $ 0.2$                               & 0.0                     & \textbf{7.5}              & 0.0                     & \textbf{21.0} \\ \hline
			\multirow{3}{*}{MIM}     & $ 0.1$                               & 49.8                    & \textbf{73.8}             & 26.0                    & \textbf{65.4}             & 0.0                     & \textbf{14.8}              & $ 0.04$                             & 0.0                     & \textbf{34.3}             & 0.0                     & \textbf{29.2} \\
			& $ 0.2$                               & 5.0                     & \textbf{54.0}             & 4.2                     & \textbf{45.4}             & 0.0                     & \textbf{6.3}              & $ 0.12$                              & 0.0                     & \textbf{32.7}             & 0.0                     & \textbf{27.6} \\
			& $ 0.3$                               & 1.5                     & \textbf{43.3}             & 2.0                     & \textbf{36.9}             & 0.0                     & \textbf{4.5}              & $ 0.2$                               & 0.0                     & \textbf{32.4}             & 0.0                     & \textbf{26.0} \\ \hline
			CW                       & $ 0.0$                                      & 42.2                    & \textbf{78.0}             & 19.5                    & \textbf{57.3}             & 0.2                     & \textbf{21.8}             & $ 0.0$                                      & 0.0                     & \textbf{30.2}             & 0.0                     & \textbf{36.2} \\ \hline \hline 
		\end{tabular}
		}
    	\end{center}
		\caption{Accuracy (\%) on K/F/MNIST, CIFAR-10 and SVHN under white-box setting. For CW, the parameter is the confidence.}
		\label{table:whitebox}
\end{table*}

Hence to prevent the prediction changes after perturbation $\boldsymbol{\epsilon}$, we should minimize $|\nabla_{\boldsymbol{x}}(\log f_y(\boldsymbol{x})-\log f_j(\boldsymbol{x}))|$. Let $a_k$ denote the logit for the $k$th class softmax output, observe that 
\begin{equation}\label{eq:fundamental}
\begin{split}
    \nabla_{\boldsymbol{x}}\log f_y - \nabla_{\boldsymbol{x}}\log f_j &= \frac{\nabla_{\boldsymbol{x}}f_y}{f_y} - \frac{\nabla_{\boldsymbol{x}}f_j}{f_j} \\&= \nabla_{\boldsymbol{x}} a_y - \nabla_{\boldsymbol{x}} a_j,
\end{split}
\end{equation}
because $\nabla_{\boldsymbol{x}}f_y = - \sum_k f_k f_y\nabla_{\boldsymbol{x}}a_k + f_y\nabla_{\boldsymbol{x}}a_y$, and the same holds for $\nabla_{\boldsymbol{x}}f_j$. We can equivalently change our objective to $\min_{\boldsymbol{\theta}} |\nabla_{\boldsymbol{x}} (a_y - a_j)|$.
We can estimate $|\nabla_{\boldsymbol{x}} (a_y(\boldsymbol{x})- a_j(\boldsymbol{x}))|$ using
\begin{equation}\label{eq:logit_estimate}
\begin{split}
    |\nabla_{\boldsymbol{x}} (a_y- a_j)| \approx  |(a_y(\boldsymbol{x})-a_j(\boldsymbol{x}))\\ - (a_y(\boldsymbol{x}+\hat{\boldsymbol{\epsilon}})-a_j(\boldsymbol{x}+\hat{\boldsymbol{\epsilon}}))|/|\hat{\boldsymbol{\epsilon}}|,
\end{split}
\end{equation}
where we denote $|\boldsymbol{\epsilon}|<|\hat{\boldsymbol{\epsilon}}| = \tau$ that upper bounds $|\boldsymbol{\epsilon}|$.
Note that an adversarial attack  tends to minimize $a_y(\boldsymbol{x}+\hat{\boldsymbol{\epsilon}})-a_j(\boldsymbol{x}+\hat{\boldsymbol{\epsilon}})$ so that $a_y(\boldsymbol{x}+\hat{\boldsymbol{\epsilon}})-a_j(\boldsymbol{x}+\hat{\boldsymbol{\epsilon}}) < a_y(\boldsymbol{x})-a_j(\boldsymbol{x})$.
And a robust model should instead prevent $a_y(\boldsymbol{x}+\boldsymbol{\epsilon})-a_j(\boldsymbol{x}+\boldsymbol{\epsilon})<0$ to ensure a correct prediction when attacked, so we have the following inequality for a robust model under attack
\begin{equation}
    0<a_y(\boldsymbol{x}+\hat{\boldsymbol{\epsilon}})-a_j(\boldsymbol{x}+\hat{\boldsymbol{\epsilon}}) < a_y(\boldsymbol{x})-a_j(\boldsymbol{x}).
\end{equation}
Then Eq.~(\ref{eq:logit_estimate}) can be upper bounded by
\begin{equation}
    |\nabla_{\boldsymbol{x}} (a_y- a_j)| < |a_y(\boldsymbol{x})-a_j(\boldsymbol{x})|/|\hat{\boldsymbol{\epsilon}}|.
\end{equation}
Substitute this inequality back to Eq.~(\ref{eq:fundamental}), we get a logit constraint condition to ensure model robustness
\begin{equation}
    |\nabla_{\boldsymbol{x}}(\log f_y - \log f_j)| < |a_y(\boldsymbol{x})-a_j(\boldsymbol{x})|/|\hat{\boldsymbol{\epsilon}}| < C,
\end{equation}
where $C$ is an arbitrary positive constant thresholding robustness, hence we can optimize PC loss subject to the above condition
\begin{equation}\label{eq:lc}
    \min_{\boldsymbol{\theta}}L_{pc}(\boldsymbol{\boldsymbol{\theta}}), \text{s.t.}|a_y(\boldsymbol{x};\boldsymbol{\theta})-a_j(\boldsymbol{x};\boldsymbol{\theta})|< C' \text{ for }\forall \boldsymbol{x},
\end{equation}
where $C' = |\hat{\boldsymbol{\epsilon}}|C$. It is equivalent to write the above minimization problem with a multiplier $\lambda$
\begin{equation}\label{eq:loss}
    \min_{\boldsymbol{\theta,\lambda}}\Big(L_{pc}(\boldsymbol{\boldsymbol{\theta}})  + \frac{\lambda}{N} \sum_{\boldsymbol{x}\in D}(d_{yj}- C')\Big),
\end{equation}
where $N$ is the number of samples, and  $D$ is the set of training samples. $\lambda$ is treated as a hyper-parameter in training, $ d_{yj} = \max(0,a_y(\boldsymbol{x};\boldsymbol{\theta})-a_j(\boldsymbol{x};\boldsymbol{\theta}))$. As shown in Eq. (\ref{eq:loss}), PC loss and logit constraints are systematically integrated and simultaneously optimized during training process to enhance model adversarial robustness. 
 
\section{Experiments}

In this section, we evaluate our proposed PC loss with logit constraints along with analysis that our method does not rely on the `gradient masking' that provides a false sense of security \cite{athalye2018obfuscated}.

\noindent{\bf Datasets and models:}  We analyze seven benchmark datasets: MNIST, KMNIST, Fashion-MNIST (FMNIST), CIFAR-10, CIFAR-100, Street-View House Numbers (SVHN), and Tiny Imagenet. We scale all pixel values to $[0,1]$ following the preprocessing procedure in \cite{mustafa2019adversarial,pang2019improving}. For gray-scale image datasets (K/F/MNIST), we use a LeNet-5 model \cite{lecun1998gradient}, and for color image datasets (CIFAR-10, CIFAR-100, SVHN, Tiny Imagenet), we use a VGG-13 model \cite{simonyan2014very}. All these models are trained using Adam optimizer with a initial learning rate of 0.01 and a batch size of 256. For our method, we first warm up the training process for $T$ epochs ($T=50$ for K/F/MNIST and $T=150$ for other datasets) using CE loss, and then train the model using our method shown in Eqs.~(\ref{eq:pcl}) and (\ref{eq:loss}) ($\xi = 0.995, \lambda = 0.05$) for another $T$ epochs whereas we directly train the baseline using CE loss for $2T$ epochs.
\begin{table}[h]
\begin{center}
    \resizebox{\linewidth}{!}{

\begin{tabular}{c|c|ccc|c|ccc}
\hline\hline
\multirow{2}{*}{Attacks} & \multirow{2}{*}{Param.} & \multicolumn{3}{c|}{MNIST}           & \multirow{2}{*}{Param.} & \multicolumn{3}{c}{CIFAR-10}        \\
                         &                         & CE   & GCE*          & Ours          &                         & CE   & GCE*          & Ours          \\ \hline\hline
\multirow{3}{*}{FGSM}    & $ 0.1$                  & 71.5 & \textbf{87.7} & 80.5          & $ 0.04$                 & 12.7 & 41.2          & \textbf{58.4} \\
                         & $ 0.2$                  & 51.6 & 62.7          & \textbf{76.3} & $ 0.12$                 & 10.3 & 14.8          & \textbf{17.3} \\
                         & $ 0.3$                  & 31.8 & 47.2          & \textbf{65.0} & $ 0.2$                  & 7.0  & 11.8          & \textbf{12.0} \\ \hline
\multirow{3}{*}{BIM}     & $ 0.1$                  & 52.8 & 61.9          & \textbf{72.0} & $ 0.04$                 & 0.0  & \textbf{19.6} & 16.6          \\
                         & $ 0.2$                  & 4.5  & 34.5          & \textbf{48.6} & $ 0.12$                 & 0.0  & 3.0           & \textbf{3.4}  \\
                         & $ 0.3$                  & 1.5  & 33.5          & \textbf{39.5} & $ 0.2$                  & 0.0  & 2.0           & \textbf{2.6}  \\ \hline
\multirow{3}{*}{PGD}     & $ 0.1$                  & 49.0 & 51.9          & \textbf{72.3} & $ 0.04$                 & 0.0  & 5.9           & \textbf{10.2} \\
                         & $ 0.2$                  & 3.3  & 9.6           & \textbf{50.2} & $ 0.12$                 & 0.0  & 1.9           & \textbf{3.5}  \\
                         & $ 0.3$                  & 0.8  & 2.2           & \textbf{39.7} & $ 0.2$                  & 0.0  & 1.6           & \textbf{2.7}  \\ \hline
\multirow{3}{*}{MIM}     & $ 0.1$                  & 49.8 & 61.2          & \textbf{73.8} & $ 0.04$                 & 0.0  & 15.4          & \textbf{16.0} \\
                         & $ 0.2$                  & 5.0  & 39.8          & \textbf{54.0} & $ 0.12$                 & 0.0  & \textbf{13.1} & 11.6          \\
                         & $ 0.3$                  & 1.5  & 38.8          & \textbf{43.3} & $ 0.2$                  & 0.0  & \textbf{12.7} & 11.2          \\ \hline
C\&W                     & 0.0                     & 0.0  & 25.6          & \textbf{30.1} & 0.0                     & 0.0  & 0.8           & \textbf{3.3}  \\ \hline\hline
\end{tabular}
}
\end{center}
\caption{Accuracy (\%) between GCE and our method on MNIST and CIFAR10 under white-box setting. *Results are directly from \cite{chen2019improving}.}
\label{table:compare}
\end{table}
\begin{table*}[t]
\begin{center}
\resizebox{0.75\linewidth}{!}{
\begin{tabular}{c|c|cc|cc|cc|c|cc|cc}
\hline \hline
\multirow{2}{*}{Attacks} & \multirow{2}{*}{Param.} & \multicolumn{2}{c|}{MNIST} & \multicolumn{2}{c|}{KMNIST} & \multicolumn{2}{c|}{FMNIST} & \multirow{2}{*}{Param.} & \multicolumn{2}{c|}{CIFAR-10} & \multicolumn{2}{c}{SVHN} \\
                         &                         & CE      & Ours             & CE       & Ours             & CE       & Ours             &                         & CE       & Ours              & CE      & Ours            \\ \hline \hline
\multirow{3}{*}{PGD}     & $ 0.1$          & 96.6    & \textbf{98.1}    & 90.8     & \textbf{92.5}    & 65.5     & \textbf{74.0}    & $ 0.04$         & 19.5     & \textbf{42.2}     & 43.6    & \textbf{48.8}   \\
                         & $ 0.2$          & 84.3    & \textbf{92.8}    & 76.7     & \textbf{85.0}    & 50.4     & \textbf{57.1}    & $ 0.12$         & 13.2     & \textbf{38.0}     & 19.5    & \textbf{28.1}   \\
                         & $ 0.3$          & 61.1    & \textbf{85.6}    & 59.1     & \textbf{77.7}    & 47.8     & \textbf{53.5}    & $ 0.2$          & 16.7     & \textbf{35.6}     & 13.2    & \textbf{24.5}   \\ \hline
\multirow{3}{*}{MIM}     & $ 0.1$          & 96.4    & \textbf{98.1}    & 90.4     & \textbf{92.2}    & 62.4     & \textbf{72.3}    & $ 0.04$         & 17.2     & \textbf{42.0}     & 40.1    & \textbf{44.2}   \\
                         & $ 0.2$          & 84.2    & \textbf{94.7}    & 74.9     & \textbf{83.0}    & 43.3     & \textbf{54.2}    & $ 0.12$         & 1.3      & \textbf{16.3}     & 13.3    & \textbf{21.7}   \\
                         & $ 0.3$          & 56.7    & \textbf{81.2}    & 48.4     & \textbf{63.8}    & 30.5     & \textbf{36.0}    & $ 0.2$          & 0.3      & \textbf{11.2}     & 10.0    & \textbf{16.2}   \\ \hline
SPSA   & $ 0.3$              & 72.9    & \textbf{95.7}    & 50.2     & \textbf{78.0}             & 4.3      & \textbf{39.8}    & $ 0.3$              & 0.0      & \textbf{45.3}     & 4.0     & \textbf{58.0}   \\
                     \hline \hline
\end{tabular}
}
\end{center}
\caption{Accuracy (\%) on K/F/MNIST, CIFAR-10 and SVHN under black-box setting.}
\label{table:black}
\end{table*}

\noindent{\bf Attack types} In the adversarial setting, there are two main threat models: white-box attacks where the adversary possesses complete knowledge of target model, including its architecture, training method and learned parameters, and black-box attacks where the adversary does not have access to the information about trained classifier but is aware of the classification task. We evaluate the robustness of our proposed method against both white-box and black-box attacks.

\begin{table}[ht]
	
	\begin{center}
	    \resizebox{\linewidth}{!}{

\begin{tabular}{c|c|ccc|ccc}
\hline\hline
\multirow{2}{*}{Attacks} & \multirow{2}{*}{Param.} & \multicolumn{3}{c|}{CIFAR-100} & \multicolumn{3}{c}{Tiny ImageNet} \\
                         &                           & CE    & GCE   & Ours           & CE      & GCE   & Ours            \\ \hline\hline
Clean                    & -                         & 40.2  & 64.5  & \textbf{67.7}  & \textbf{38.2}    & 32.8  & 37.7            \\ \hline
\multirow{3}{*}{PGD}     & $0.005$                   & 11.4  & 24.4  & \textbf{56.7}  & 11.3    & 8.9   & \textbf{24.8}   \\
                         & $0.010$                   & 2.0   & 14.8  & \textbf{54.6}  & 2.8     & 2.6   & \textbf{19.2}   \\
                         & $0.015$                   & 0.4   & 9.1   & \textbf{52.6}  & 0.8     & 1.0   & \textbf{15.8}   \\ \hline
\multirow{3}{*}{MIM}     & $0.005$                   & 8.7   & 21.9  & \textbf{55.8}  & 7.9     & 7.3   & \textbf{23.5}   \\
                         & $0.010$                   & 1.4   & 12.3  & \textbf{52.8}  & 1.9     & 2.1   & \textbf{18.0}   \\
                         & $0.015$                   & 0.3   & 7.4   & \textbf{48.9}  & 0.6     & 1.4   & \textbf{14.6}   \\ \hline
SPSA                     & $0.015$                   & 3.9   & 11.5  & \textbf{22.1}  & 6.3     & 7.3   & \textbf{16.2}   \\ \hline\hline
\end{tabular}
}
	\end{center}
	
	\caption{Accuracy (\%) on CIFAR-100 and Tiny ImageNet between CE loss, GCE and our new PC loss.}
	\label{table:cifar100}
\end{table}
\subsection{Results}
\noindent\textbf{Performance on white-box attacks} Following the attack settings in \cite{chen2019improving}, we crafted adversarial examples in a non-targeted way with respect to allowed perturbation $\epsilon$ for gradient-based attacks, i.e., FGSM, BIM, PGD and MIM. The number of iterations is set to 10 for BIM and 40 for MIM and PGD while perturbation of each step is 0.01. For parameters of optimization-based attack C\&W, the maximum iteration steps are set to 100, with a learning rate of 0.001, and the confidence is set to 0. 

The results (Table \ref{table:whitebox}) demonstrate that our proposed PC loss with logit constraints outperforms the CE loss under white-box attacks while maintaining the comparable level of performance on the clean image classification. The improvement is even more significant on stronger attacks. 

Besides comparing to the standard CE loss, we also compare our defense approach with a closely related Guided Complement Entropy (GCE) approach \cite{chen2019improving}. To ensure a fair comparison we use the exactly same models (LeNet-5 for MNIST and ResNet56 for CIFAR-10) and parameters (max iterations of C\&W is 1000) as in the GCE paper. In Table~\ref{table:compare}, it is evident that our method outperforms GCE in the vast majority of settings.

\noindent\textbf{Performance on black-box attacks} The performance under black-box setting is critical to substantiate adversarial robustness since it is  closer to the real-world scenario where an adversary has no access to the trained classifier. During inference time, black-box adversary uses a substitute model trained on the same dataset to generate adversarial samples to attack the target model. In our cases, we use a 3-layer CNN as the substitute model for LeNet-5 and ResNet-56 for VGG-13 to generate black-box attacks. Similar to \cite{pang2019improving}, we adopt PGD and MIM, the two most commonly used attack methods under the black-box setting. We then further evaluate our defense method using a gradient-free attack approach, i.e., SPSA, as in \cite{carlini2019evaluating}, which performs numerical approximation on the gradients using test data. The learning rate of SPSA is set to 0.01, and the step size is $\delta = 0.01$ \cite{uesato2018adversarial}. As shown in Table \ref{table:black}, the model trained with our PC loss improves robustness against the black-box attacks. 

\noindent\textbf{Larger-scale experiments on CIFAR-100 and Tiny ImageNet} We also evaluate our method on larger and more complex CIFAR-100 and Tiny ImageNet datasets under both white-box attacks (PGD, MIM) and black-box attack (SPSA). Similar to \cite{pang2019improving}, we reduce the perturbation budget to the range of [0.005, 0.015] and attack iterations to 10 due to the increased data complexity and scale. As shown in Table \ref{table:cifar100}, our method significantly improves the model's adversarial robustness compared to the CE loss and GCE while  maintaining the  comparable level of performance on the clean image classification. Recall our observation that the most probable false classes are more vulnerable to attacks. GCE flattens the probabilities on false classes and thus enlarges the gap between true class and the most probable false class to increase model's robustness. However, when dataset become complex with more classes, this gap is smaller due to generally lower output probability for the true class, resulting a limited robustness improvement. On the other hand, our method directly maximizes the probability gap and thus is more suitable for large scale datasets.

\begin{table*}[t]
\begin{center}
\resizebox{\linewidth}{!}{
\begin{tabular}{c|c|cccc|cccc|cccc|c|cc}
\hline\hline
\multirow{2}{*}{Attacks} & \multirow{2}{*}{Param.} & \multicolumn{4}{c|}{MNIST}            & \multicolumn{4}{c|}{KMNIST}           & \multicolumn{4}{c|}{FMNIST}       & \multirow{2}{*}{Param.}     & \multicolumn{2}{c}{CIFAR-10} \\
                         &                         & CE+AT & GCE+AT & Ours+AT       & Ours & CE+AT & GCE+AT & Ours+AT       & Ours & CE+AT & CGE+AT & Ours+AT       & Ours &  & CE+AT & Ours+AT \\ \hline\hline
BIM                      & $ 0.3$                  & 27.0  & 28.3   & \textbf{86.1} & 39.5 & 58.4  & 8.1    & \textbf{65.9} & 60.5 & 0.1   & 4.8    & \textbf{18.6} & 4.9  &0.04 & 17.3  &  \textbf{38.7}\\
PGD                      & $ 0.3$                  & 3.2   & 26.8   & \textbf{72.3} & 39.7 & 37.2  & 0.5    & \textbf{48.7} & 33.2 & 0.0   & 2.7    & \textbf{13.4} & 2.8  &0.04  & 11.4 &  \textbf{33.7}\\
MIM                      & $ 0.3$                  & 10.8  & 27.7   & \textbf{78.8} & 43.3 & 22.3  & 15.7   & \textbf{51.9} & 36.9 & 0.0   & 1.4    & \textbf{9.1} & 2.2  &0.04  & 9.0  &  \textbf{33.3}\\
C\&W                     & $0.0$                   & 75.5  & 69.4   & \textbf{96.4} & 78.0 & 47.9  & 48.2   & \textbf{67.1} & 57.3 & 4.0   & 21.5   & \textbf{29.8} & 21.8 &0.0  & 0.0  &  \textbf{33.7}\\
SPSA                     & $ 0.3$                  & 77.3  & 56.6   & \textbf{97.1} & 95.7 & 72.0  & 69.9   & \textbf{79.0} & 78.0 & 14.2  & 30.0   & \textbf{41.6} & 39.8 &0.3  & 6.5  &  \textbf{39.6}\\ \hline\hline
\end{tabular}
}
\end{center}

\caption{Accuracy (\%) on K/F/MNIST and CIFAR-10 with adversarial training under both white- and black-box attacks.}
\label{table:AT}
\end{table*}
\begin{figure*}[t]
\centering
\includegraphics[width=\linewidth]{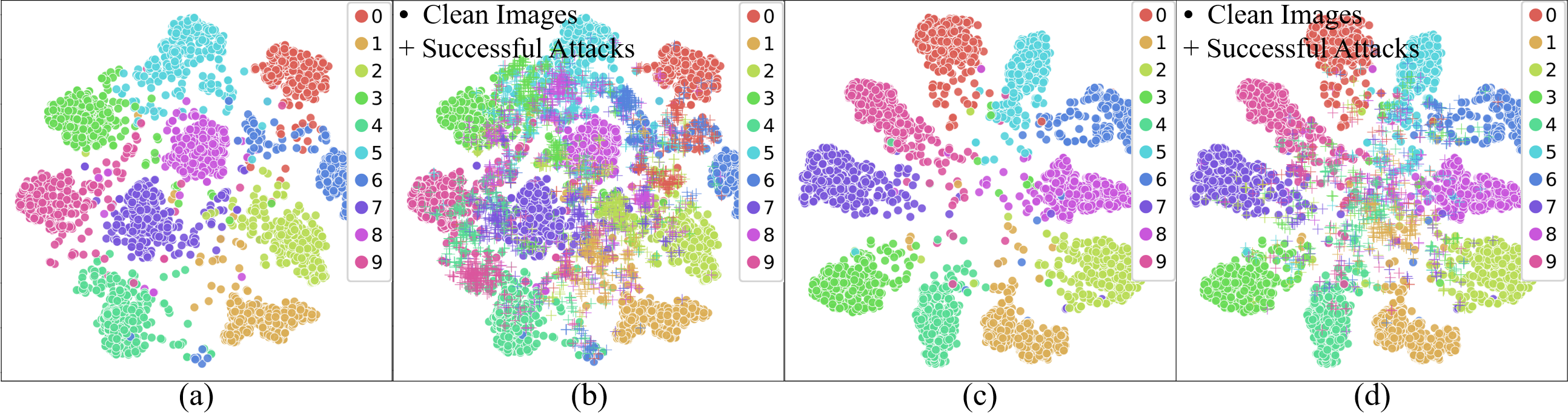}
\caption{T-SNE visualization of the penultimate layer of the model trained by CE loss (a,b) and our PC loss (c,d) on MNIST dataset. (a,c) display only clean images whereas (b,d) also include successful attacks generated with FGSM ($\epsilon = 0.3$).}
\label{fig:mnist_vis}
\end{figure*}

\noindent\textbf{Combining with adversarial training}
To demonstrate our method's compatibility and synergy with other adversarial defense techniques, we investigate the performance of our method in combination with adversarial training. Our goal is not to beat adversarial training, instead we attempt to show our method can be combined with it to further improve adversarial robustness. During training, we augment the dataset with adversarial samples generated using FGSM with perturbation range of $[0.1, 0.3]$ for gray-scale image datasets (K/F/MNIST), and 5-step PGD with perturbation range of $[0.0, 0.1]$ for color image dataset (CIFAR-10). The ratio of adversarial examples and clean images in each training mini-batch remains $1:1$. For gray-scale image datasets, table \ref{table:AT} shows that integrating our method with adversarial training further improves the model's adversarial robustness under both white-box (PGD, BIM, MIM, CW) and black-box attack settings (SPSA). Furthermore, our PC loss with adversarial training outperforms GCE with adversarial training, which demonstrates our method has better compatibility with other defense techniques. It is worth mentioning that our method alone outperforms the fast version adversarial training on gray-scale datasets, which generates adversarial training examples by one-step FGSM attack. And for color image dataset (CIFAR-10), we augment the dataset with adversarial examples crafted by more advanced PGD attacks. The result shows the same trend, and the performance gain is more pronounced on this more challenging dataset.

\noindent\textbf{Feature Space Visualization} In order to visually dissect the advantages of PC loss over the CE loss, we also inspect the feature space of trained models using t-SNE on MNIST datasets. As shown in Figure~\ref{fig:mnist_vis}a, the model trained with CE loss has a large portion of clean images lay across the boundaries between different classes thus easily to be manipulated to become adversarial samples. On the contrary, for the model trained with our PC loss with logit constraints, in Figure~\ref{fig:mnist_vis}c, the samples of each class has clear boundaries and are evenly distributed around the center with a minimal overlap. Note that the samples locate near the center are `hard samples' for a classifier even without adversarial attacks.

Looking into the successful attacks (labeled with `+') in Figure~\ref{fig:mnist_vis}, we find the predictive behavior of CNN on adversarial samples is consistent to our hypothesis. In Figure~\ref{fig:mnist_vis}b, for a model (LeNet-5) trained with CE loss, adversarial samples are mostly located to the nearest classes corresponding to the most probable false classes. For example, many adversarial attacks generated based on class 5 are located within the class 8 of clean images and {\it vice versa}. While in Figure~\ref{fig:mnist_vis}d, for a model trained with PC loss, due to the large margin between classes, the adversarial samples are harder to cross the boundaries with the only exception that the adversarial samples are distributed near the center of the feature space where hard samples are usually located.

\subsection{Identifying Gradient Masking}

Previous defense strategies \cite{buckman2018thermometer,xie2017mitigating} rely on the effect of gradient masking, which was considered as a false sense of security \cite{athalye2018obfuscated}. Briefly, these defenses deteriorate the gradient information to make gradient-based attack methods hard to generate effective adversarial examples. However, these defenses can be easily defeated by black-box or gradient-free attackers. We show that our method does not rely on gradient masking on the basis of characteristics defined in \cite{athalye2018obfuscated,carlini2019evaluating}. (1) Iterative attacks have better performance than one-step attack: Our results in Table \ref{table:whitebox} indicate that the iteration-based attacks (BIM, MIM, PGD) are more successful in generating adversarial attacks than single step method (FGSM). (2) Robustness against Black-box attacks is higher than white-box attacks: When model's gradients information is manipulated by the defender, the attacker can recover the gradient with black-box attacks and perform more successful attacks than using white-box attacks \cite{papernot2017practical}. However, the results in Tables \ref{table:whitebox} \& \ref{table:black} demonstrate that our method is more effective against black-box attacks and thus does not obfuscate gradients. (3) Increasing perturbation budget will increase attack success: As shown in the Table \ref{table:whitebox}, increase of perturbations monotonically enhances the attacks. With a large budget (e.g., $\epsilon$ = 0.3), the attack success rate is close to 100\%.

\section{Conclusion}
We propose a novel PC loss with logit constraints inspired by the predictive behavior of CNN on adversarial samples. A CNN trained with our PC loss can achieve impressive robustness against adversarial samples without compromising performance on clean images nor requires additional procedures/computing, making it scalable to large-scale datasets. In addition, our PC loss is flexible and compatible with other defense methods, e.g.,  as a drop-in replacement of CE loss to supervise adversarial training. In future work, we plan to extensively investigate the connection of predictions between adversarial and clean samples in more general settings.







\bibliographystyle{aaai}
\bibliography{egbib}

\end{document}